\documentclass[10pt,conference]{IEEEtran}
\IEEEoverridecommandlockouts
\usepackage{cite}
\usepackage{amsmath,amssymb,amsfonts}
\usepackage{graphicx}
\usepackage{textcomp}
\usepackage{xcolor}
\def\BibTeX{{\rm B\kern-.05em{\sc i\kern-.025em b}\kern-.08em
    T\kern-.1667em\lower.7ex\hbox{E}\kern-.125emX}}
    
\usepackage{multirow}
\usepackage{adjustbox}
\usepackage{booktabs}
\usepackage{ragged2e}
\usepackage{tabularx}
\usepackage{caption}
\usepackage{subcaption}
\usepackage{algorithm} 
\usepackage{algpseudocode} 
\usepackage{csquotes}
\usepackage{verbatim}
\usepackage{color}
\usepackage{soul}
\usepackage[most]{tcolorbox}
\usepackage{flushend}
\newcommand{\hlcyan}[1]{{\sethlcolor{cyan}\hl{#1}}}

\begin{document}

\title{Collaboration among Multiple Large Language Models for Medical Question Answering\\
}

\author{\IEEEauthorblockN{1\textsuperscript{st} Kexin Shang}
\IEEEauthorblockA{\textit{College of Computing \& Informatics} \\
\textit{Drexel University}\\
Philadelphia, USA \\
ks4252@drexel.edu}
\and
\IEEEauthorblockN{2\textsuperscript{nd} Chia-Hsuan Chang}
\IEEEauthorblockA{\textit{College of Computing \& Informatics} \\
\textit{Drexel University}\\
Philadelphia, USA \\
shane.chang.tw@gmail.com}
\and
\IEEEauthorblockN{3\textsuperscript{rd} Christopher C. Yang}
\IEEEauthorblockA{\textit{College of Computing \& Informatics} \\
\textit{Drexel University}\\
Philadelphia, USA \\
chris.yang@drexel.edu}
}

\maketitle

\begin{abstract}
Empowered by vast internal knowledge reservoir, the new generation of large language models (LLMs) demonstrate untapped potential to tackle medical tasks. However, there is insufficient effort made towards summoning up a synergic effect from multiple LLMs' expertise and background. In this study, we propose a multi-LLM collaboration framework tailored on a medical multiple-choice questions dataset. Through post-hoc analysis on 3 pre-trained LLM participants, our framework is proved to boost all LLMs reasoning ability as well as alleviate their divergence among questions. We also measure an LLM's confidence when it confronts with adversary opinions from other LLMs and observe a concurrence between LLM's confidence and prediction accuracy. 

\end{abstract}


\section{Introduction}
The rapid evolution of generative AI and specifically large language models (LLMs) has spurred interest in adoption of these technologies in various fields. In the medical domain, several applications of LLMs have been studied extensively, including medical question answering (QA), clinical decision support, information extraction from clinical notes, and more. 

In addition to large general purpose LLMs such as OpenAI's GPT, Google's Gemini, Meta's Llama, and Anthropic's Claude, multiple domain-specific LLMs have emerged, fine-tuned for specific tasks. For clinical tasks, these include \textit{Med42-70B} \cite{christophe2024med42evaluatingfinetuning}, \textit{ClinicalCamel-70B} \cite{toma2023clinicalcamelopenexpertlevel}, \textit{Meditron-70B} \cite{chen2023meditron70b}, and \textit{MedAlpaca-13B} \cite{han2023medalpacaopensourcecollection}, among others. While some of these LLMs may have overlap in their fine-tuning datasets, they adopt distinct data preprocessing and training methods and vary in both size and architecture. Different studies have applied these various LLMs to the task of medical QA and compared their performance, but few have adopted multi-LLM approaches for this task. Additionally, some studies have proposed different mechanisms to enhance or improve the reasoning of LLMs for medical QA, but these improvements are typically based on the generations of the same LLM. Given the critical role of clinical reasoning in medical tasks, and the variable performance and reasoning of different LLMs, there is potential to explore whether multi-LLM collaboration for medical QA can reduce errors and improve performance. 

To explore potential collaboration approach to enhance the performance of multiple LLMs in a medical QA dataset USMLE, we design and evaluate a framework where LLMs work collaboratively by sharing their reasonings on a medical question. Our results validate that this process of collaboration efficiently mitigate the disagreement among LLMs as well as leading to individual and overall improved performance in the task. Besides, in the setting of conflicting answers from the other LLMs, we consider an LLM's tendency to ``concede'' (change its answer) or to ``insist'' (not change its answer) as a measure of its inherent ``confidence'', observing a positive pattern between LLM's reasoning capability and its confidence degree. Lastly, we observe that all LLMs show a gap in consistency between questions they answered correctly and incorrectly.

\section{Related Work}
\subsection{Medical QA with LLMs}

Multiple studies have evaluated and reported on the use of LLMs in medical QA using different LLMs and datasets.  These include medical textbook questions on ChatGPT \cite{kumah-crystal_chatgpt_2023}, medical licensing exam questions on ChatGPT, GPT-3, and GPT-4 \cite{kung_performance_2023, gilson_how_2023, nori_capabilities_2023, thirunavukarasu_trialling_2023}, and real world patient questions on ChatGPT and on Google Bard \cite{chowdhury_can_2023, lim_benchmarking_2023}.  These studies have primarily focused on evaluating the performance of specific LLMs with some additionally analyzing the reasoning provided. 

Lucas \textit{et al.} \cite{lucas_reasoning_2024} proposed a mechanism to improve the reasoning and consistency of LLMs in medical QA using an ensemble reasoning approach. However, this method, while promising, also involves LLMs reviewing their own reasonings, which can cause LLM hallucinations and is susceptible to confirmation bias \cite{huang_large_2024, xie_adaptive_2024}. Additionally, it is computationally demanding, requiring multiple calls to the API. Multi-LLM collaborations on medical QA have not been extensively explored as a mechanism for improved reasoning and performance.

\subsection{LLM collaboration}
Collaboration is commonly understood as the act of working together to complete a given task or achieve a common goal.  In the context of LLMs, the concept of collaboration can take on different forms. 
Feng \textit{et al.} \cite{feng_dont_nodate} propose two multi-LLM collaboration-based pathways: ``cooperate" and ``compete" to identify and mitigate knowledge gaps in LLMs.  They enable ``abstention", where an LLM, after interacting with other LLMs, should abstain from answering a question incorrectly.  In the cooperation setting, a ``judge" LLM will compare feedback from other LLMs to determine if a particular LLM should abstain from answering. In the compete setting, an LLM is challenged by others with alternative answers and if on average it is ``swayed" by these conflicting answers then it should abstain. Through this abstention mechanism, their multi-LLM collaboration has the potential to mitigate hallucinations. 

Zhang \textit{et al.} \cite{zhang-etal-2024-exploring} take a psycho-social view to exploring LLM collaboration. They conceptualize LLM agents as having the traits of either being ``easy-going" or ``overconfident" and craft prompts to enable these traits. They then define two ``thinking patterns" in multi-round LLM conversations: ``debate" and ``reflection" and evaluate different collaborative strategies by simulating a ``machine society" comprising different LLM agents and permuting the two thinking patterns.

Fang \textit{et al.} \cite{fang2024counterfactualdebatingpresetstances} propose ``CounterFactual Multi-Agent Debate (CFMAD)", a framework where a ``critic" LLM is crafted to always hold counterfactual viewpoints to the ``assistant" LLM. These two LLMs debate on every possible choice option of a multiple-choice question, and their debate transcript is subsequently examined by a third-party ``judge" LLM to decide the final answer. Their experiment results suggest that CFMAD helps reduce or eliminate hallucinations in LLMs.

In the domain of medical QA, Yang \textit{et al.} \cite{yang_one_2023} investigate an ensemble approach, LLM-Synergy, comparing two different ensembling methods and finding improvements in accuracy. While their method does not involve the constituent LLMs exchanging information, they leverage the diversity and capabilities of different LLMs for medical QA. However, their approach does not incorporate the LLMs reasoning.

The highlighted studies propose and evaluate various forms of multi-LLM collaboration, using differing mechanisms.  However, we note two gaps.  While some of the collaboration mechanisms involve debate or exchanging opinions on a question, they may not use the chain-of-thought \cite{wei2023chainofthoughtpromptingelicitsreasoning, kojima2023largelanguagemodelszeroshot} approach to obtain the LLM reasoning for the generated answer, which has been demonstrated to improve model performance.  Additionally, there is a lack of studies specifically looking at medical QA where clinical reasoning is a crucial part of any decision making and question answering.  

\section{Methods and Materials}

\subsection{Data Source and Preparation}
We use a medical question dataset from the sample exams of the United States Medical Licensing Examination (USMLE). This exam contains three steps that assess medical professionals' foundational knowledge across medical science, clinical medicine, biomedical science and other domain skills \cite{ghaffari-rafi_multivariable_2019}. Each step has different emphasis:

\begin{itemize}
    \item Step 1 assesses important concepts of the sciences basic to the practice of medicine.
    \item Step 2 assesses an examinee’s ability to apply medical knowledge, skills, and understanding of clinical science essential for the provision of patient care under supervision.
    \item Step 3 assesses the medical knowledge, understanding of biomedical and clinical science essential for the unsupervised practice of medicine.
\end{itemize}

Unlike the actual USMLE exam, the sample exams only comprise multiple-choice questions, providing a comprehensive dataset to serve our purpose of multi-LLM collaboration study in medical QA. Since all LLMs used in this study are text-only models, we follow the data processing proposed in the previous work~\cite{lucas_reasoning_2024} to filter out questions with complex structure such as having tables, charts, text-based patient records or pictures. After streamlining our dataset, the distribution of questions is shown in Table \ref{dataset_count}.

\begin{table}[t]
\centering
\caption{Sample Size of USMLE Sub-dataset}
\begin{tabular}{@{}lccc@{}}
\toprule
\multicolumn{4}{c}{USMLE Sample Questions}     \\ \midrule
                    & Step 1 & Step 2 & Step 3 \\
Number of questions & 87     & 100    & 118    \\ \bottomrule
\end{tabular}
\label{dataset_count}
\end{table}

\begin{algorithm} 
	\caption{Zero-shot Chain-of-Thought with Self-Consistency (ZS-CoT-SC)}
        \label{alg. zscot_sc}
        
        \begin{algorithmic}[1]
        \Require a question $q$, an LLM $\phi$, a repeat number $n$, a summarizer $\psi$
        \For {$i \in \{1,n\}$} \Comment{Apply self-consistency}
            \State $r_i = \phi(T_{reasoning}(q))$
            \State $y_i = \phi(T_{answer}(q, r_i))$
        \EndFor
        \State $\hat{y}_{q,\phi} = \text{MajorityVote}(\{y_i|1 \leq i \leq n\})$ \Comment{Pick the most consistent answer} 
        \State $\hat{\textbf{r}}_{q,\phi} = \{r_i | 1 \leq i \leq n, y_i = \hat{y}_{q,\phi}\}$ 
        \Comment{Collect reasonings associated with the majority vote} 
        \State $\hat{s}_{q,\phi} = \psi(T_{summary}(\hat{\textbf{r}}_{q,\phi}))$ \Comment{Summarize the list of reasonings}
        \State \Return $R_{q,\phi} = (\hat{y}_{q,\phi},\hat{s}_{q,\phi})$
	\end{algorithmic} 
\end{algorithm}

\begin{algorithm} 
	\caption{Collaboration Procedure}
        \label{alg. collaboration procedure}
        
        \begin{algorithmic}[1]
        \Require a set of disagreed questions $Q^{dis}$, a set of LLMs $\Phi$, a summarizer $\psi$, initial predictions $\{R_{q,\phi}|q \in Q^{dis}, \phi \in \Phi\}$
        \While{$P^{con} \leq 80\%$}
            \For{$q \in Q^{dis}$}
                \State $Trans_q = \{R_{q,\phi} | \phi \in \Phi \}$ 
                \For{$\phi \in \Phi$}
                    \State $R_{q,\phi} = \text{ZS-COT-SC}(q+Trans_q, \phi, \psi)$ \Comment{Re-apply Algorithm 1 with a different prompt $T_{reasoning\_review}$}
                \EndFor
            \EndFor
            \State Update $Q^{dis}$ and $P^{con}$ based on the updated predictions
        \EndWhile
	\end{algorithmic} 
\end{algorithm}

\subsection{Iterative Collaboration Framework (ICF)}

We propose a multi-LLM Iterative Collaboration Framework (ICF) to facilitate the information exchange of multiple-choice questions among LLMs. As shown in Fig. \ref{diagram}, ICF consists of two parts: Zero-shot Chain-of-Thought with Self-consistency (ZS-CoT-SC) and Collaboration Loop. 

\subsubsection{Zero-shot Chain-of-Thought with Self-consistency (ZS-CoT-SC)}

\begin{figure*}[ht]
    \centering
    \includegraphics[width=1.0\textwidth]{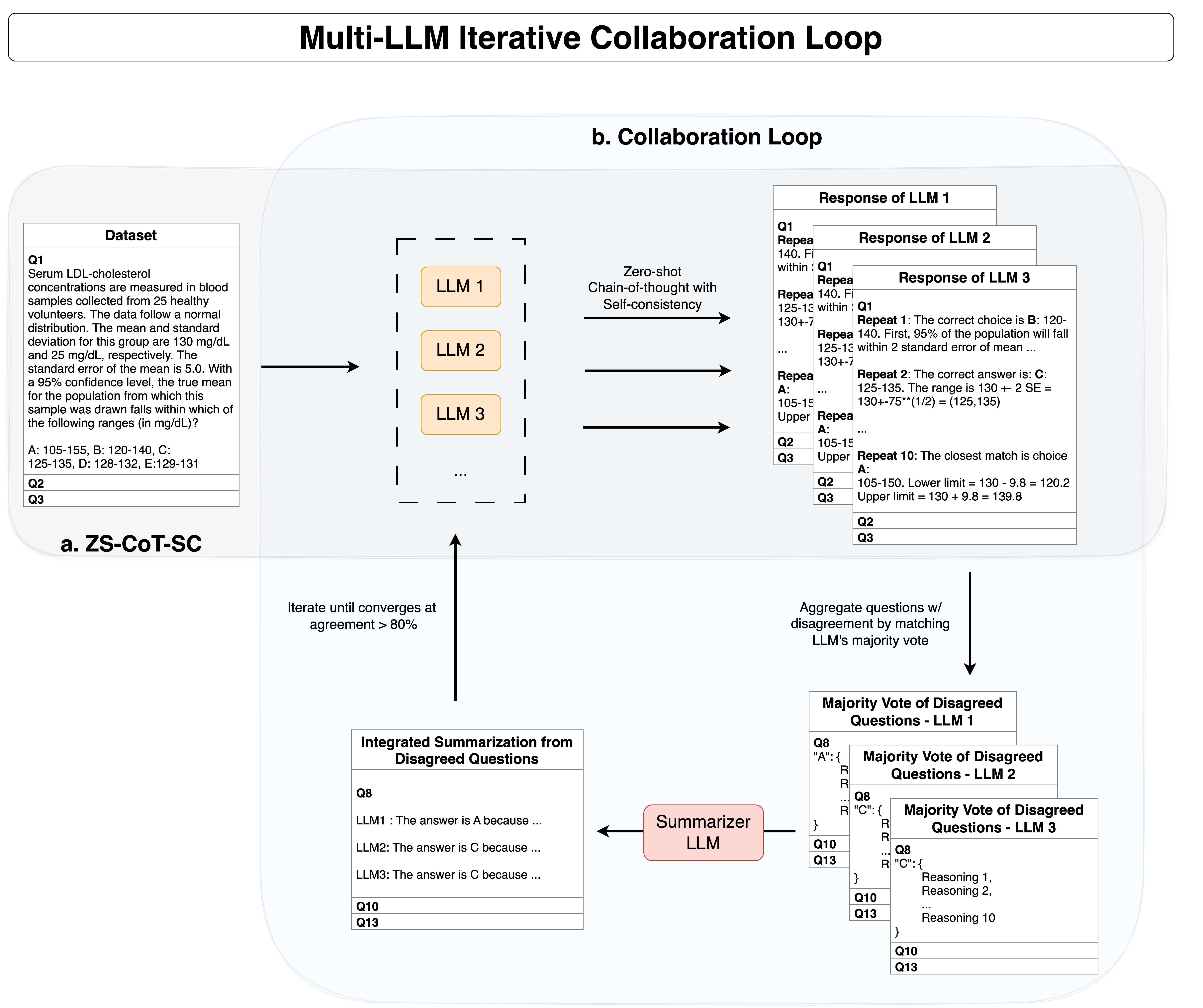}
    \caption{ICF Framework consisted of two parts: (a) ZS-CoT-SC and (b) Collaboration Loop}
    \label{diagram}
\end{figure*}

At the first step of ICF, we apply self-consistency decoding strategy and zero-shot chain-of-thought prompting \cite{kojima2023largelanguagemodelszeroshot} (ZS-CoT) to LLM inference. We define this process as ``\textbf{Z}ero-\textbf{s}hot \textbf{C}hain-of-\textbf{T}hought with \textbf{S}elf-\textbf{C}onsistency (ZS-CoT-SC)" (Fig.\ref{diagram} a.). Its pseudocode is shown in Algorithm \ref{alg. zscot_sc}. 

Denoting $Q$ a collection of questions, for each question ${q} \in Q$, we use a base LLM $\phi$ to generate $n=10$ responses (i.e., line 1 - line 3). During this process, ZS-CoT \cite{kojima2023largelanguagemodelszeroshot} is utilized to first elicit a step-wise reasoning with prompt template $T_{reasoning}$ and then a letter choice with prompt template $T_{answer}$. Next, the most consistent letter choice out of $n$ generations will be selected as the majority vote $\hat{y}_{q,\phi}$ for $q$ given $\phi$ (i.e., line 5). After identifying $\hat{y}_{q,\phi}$, all of its associated reasonings will be extracted and concatenated in line 6. To reduce the information redundancy of the concatenated reasonings, they are summarized into one paragraph, $\hat{s}_{q,\phi}$, by an external summarizer LLM $\psi$ with prompt template $T_{summary}$ (i.e., line 7). At the end of ZS-CoT-SC step, each question $q$ is paired with a consolidated context $R_{q,\phi}$ containing its majority vote $\hat{y}_{q,\phi}$ and the summary of its reasoings $\hat{s}_{q,\phi}$ (i.e., line 8). The prompt templates $T_{reasoning}$, $T_{answer}$, and $T_{summary}$ can be found in Fig. \ref{prompt_example} and \ref{prompt_summarizer}.

Having collected results of $Q$ across all LLM participants in ZS-CoT-SC step, we determine if LLMs have \textbf{consensus} or \textbf{disagreement} on $q$ by aligning their majority votes: if all LLMs propose the same majority vote towards, they are said to reach \textbf{consensus} on $q$; else, if at least one LLM proposes a different majority vote than others, they are said to have \textbf{disagreement} on $q$. Following this process, $Q$ is further classified into two subsets after ZS-CoT-SC:

\begin{itemize}
   \item $Q^{con}$: Question set where all LLMs have consensus
   \item $Q^{dis}$: Question set where LLMs have disagreement
\end{itemize}

On the top of this classification scheme, we denote the \textbf{consensus rate} $P^{con}$ across all LLMs as:

\begin{equation} \label{eq:2}
P^{con} = \frac{|Q^{con}|}{|Q|} * 100\%
\end{equation} 

$P^{con}$ is used as a termination criterion to decide if team members are diverging enough on dataset $Q$ to necessitate an opinion exchange in the later collaboration loop (Fig.\ref{diagram}, b). We set $P^{con} \ge 80\% $ at this point to control the iteration of collaboration loop appropriately: if LLMs fail to reach at least 80\% consensus rate at the end of ZS-CoT-SC, a collaboration loop will kick off. \\

\subsubsection{Collaboration Loop} \label{collaboration_loop_method}
In the next procedure, collaboration loop (Algorithm \ref{alg. collaboration procedure}), LLMs resolve their divergence on all disagreed questions by exchanging their initial prediction generated from the preceding ZS-CoT-SC step. For each disagreed question $q \in Q^{dis}$, we aggregate predictions across all LLMs and their associated summarized reasoning into an integrated transcript $Trans_q$ (i.e., line 3). This co-produced transcript allows the entire team share expansive reasoning pathways from each other.

Then ZS-CoT-SC is reused to have each base model reviewed each disagreed question and the corresponding transcript. Unlike using two sequential prompts $T_{reasoning}$ and $T_{answer}$ in the initial ZS-CoT-SC, we use a single prompt template $T_{reasoning\_review}$ (Fig. \ref{prompt_example}) for ZS-CoT-SC to guide LLMs to critically examine reasonings from all team members and re-decide their answers (i.e., line 5). 

As an LLM's new answer might be or not be the same, we update the $Q^{dis}$ and re-calculate new consensus percentage $P^{con}$ (i.e., line 8). This collaboration loop will continue to circulate $Q^{dis}$ among LLMs until $P^{con}$ meets the termination criterion. 

\begin{figure}
    \begin{tcbraster} [raster columns=1,raster equal height, nobeforeafter]
      \begin{tcolorbox}[title=ZS-CoT-SC Prompt]
        
        \hrulefill  \hspace{0.1cm} $T_{reasoning}$ \hspace{0.1cm} \hrulefill
        
        \texttt{\textless|\text{System}|\textgreater}: You are an expert medical professional who helps to reason multiple choice questions.
        
        \hfill 
        
        \texttt{\textless|\text{Question}|\textgreater}: \{question\} \{choices\} \
    
        \hfill 
        
        \texttt{\textless|\text{Assistant}|\textgreater}: Let us think step by step. First,
    
        \hfill 
        
        \hl{Step-wise reasoning from base LLM\\...\\...}
    
        \hfill 
        
        \hrulefill \hspace{0.1cm} $T_{answer}$ \hspace{0.1cm} \hrulefill
    
        Therefore, among choice \{first letter\} through \{last letter\}, the answer (letter) is:

        \hfill 
        
        \hl{A}
    
        \hfill 
        
      \end{tcolorbox}
      \begin{tcolorbox}[title=Collaboration Loop Prompt]
      
        \hrulefill  \hspace{0.1cm} $T_{reasoning\_review}$ \hspace{0.1cm} \hrulefill
        
        \texttt{\textless|\text{System}|\textgreater}: You are a medical professional who helps resolve disagreements among other experts regarding a medical question by critically reviewing their reasoning.
        
        \hfill 
            
        \texttt{\textless|\text{Prompter}|\textgreater}:  \{question\} \{choices\} \{answers and reasonings integrated from all LLMs\} 

        \hfill        
        
        After reviewing experts' choices and reasonings, which do you agree with and why? Please first output the answer (letter) and then your reasoning:    
        
        \hfill 
        
        \texttt{\textless|\text{Assistant}|\textgreater}: 
        
      \end{tcolorbox}
    \end{tcbraster}
    \caption {Prompt templates of Med42 in ZS-CoT-SC and collaboration loop of ICF.
    (Upper) In ZS-CoT-SC, two sequential prompts $T_{reasoning}$ and $T_{answer}$ jointly formulate the basic ZS-CoT template, where the highlighted part is base LLM's response.  Then each ZS-CoT template is applied 10 times on a question via self-consistency.
    (Lower) Collaboration loop applies $T_{reasoning\_review}$ for every recursive ZS-CoT-SC, presenting all LLM's reasoning pathways and ask for re-decision at the same time.}
\label{prompt_example}
\end{figure}

\begin{figure}
    \begin{tcolorbox} [title= Summarization Prompt]
        \hrulefill  \hspace{0.1cm} $T_{summary}$ \hspace{0.1cm} \hrulefill
        
        \texttt{\textless\text{s}\textgreater}[INST] You are an expert medical professional who helps to summarize opinions from a panel of experts regarding a medical multiple choice question.
        
        \hfill 

        \hlcyan{Majority answers and aggregated 10 reasonings from one base LLM}
        
        \hfill 
        
        Please read all the response above. Then extractively summarize their opinions into one paragraph.
        
        \hfill        
        
        [/INST]Summary:
        
    \end{tcolorbox}
    \caption {Prompt templates of summerizer LLM $\psi$. In this experiment, we deploy Mixtral to summarize repetitive reasoning pathways from self-consistency sampling. The highlighted part is a integrated context of the majority vote and $n=10$ generated reasonings from one LLM participant in ICF.}
\label{prompt_summarizer}
\end{figure}

\subsection{LLM Participants}
It is our purpose to construct an LLM team with diversity, hoping every model can leverage its expertise and fill up the knowledge gap of its teammates in various medical fields. Therefore, three open-source LLMs from unique backgrounds are selected to assemble a multi-LLM system: \textit{med42-70B-v1} \cite{christophe2024med42evaluatingfinetuning}, \textit{ClinicalCamel-70B} \cite{toma2023clinicalcamelopenexpertlevel}, and \textit{Mixtral 8x7B} \cite{jiang2024mixtralexperts}. For brevity, we will refer to their short name, Med42, ClinCamel, and Mixtral in the rest of this paper. 

The first model, Med42, has 70B parameters and is specifically refined from \textit{Llama-2-70B} \cite{touvron2023llama2openfoundation} architecture for medical tasks. It is fine-tuned with 411,064 medical samples and 295,649 general samples arranging from clinical articles, dialogue, medical professional tests to clinical reports and so on. Another medical specialized LLM, ClinCamel has the same parameter size of 70B and uses the same base architecture \textit{Llama-2-70B} with difference that it is fined-tuned from an extensive data collection of clinical dialogue and utilizes QLoRA technique \cite{dettmers2023qloraefficientfinetuningquantized} during fine-tune. Mixtral is a general purpose mixture of experts (MoE) model with 46.7B parameters. In addition to demonstrating strong performance in multilingual understanding tasks, Mixtral has a 10240-token long context window, which makes it an ideal LLM to retrieve and summarize information. Thus, we create two separate instances of Mixtral for different purposes: one is a team member in ICF and the other is an external helper agent $\psi$ who summarizes opinions from other two teammates (line 7, Algorithm 1). 

All three models are reported to either match or surpass the performance of OpenAI's GPT-3.5 LLM on most benchmarks \cite{christophe2024med42evaluatingfinetuning, toma2023clinicalcamelopenexpertlevel, jiang2024mixtralexperts}. 

As LLMs can be very sensitive to prompt formatting \cite{sclar_quantifying_2023}, we tailor each model the most suitable prompts for zero-shot and collaboration inference through testing a variation of prompting formats on randomly selected 30 sample questions. The best prompts gain highest accuracy and is thus selected as template for this experiment. We set the temperature at 1 for all LLMs and retain all other parameters at their default values.


We quantize three LLMs to int8 precision to save computational resource and conduct our experiment on two NVIDIA A40 GPUs for Mixtral inference and four NVIDIA A40 GPUs for Med42 and ClinCamel inference.

\subsection{Metrics}

\subsubsection{Accuracy}

We use prediction accuracy as a metric of LLM's reasoning capacity because it is the major metric for the medical Q\&A task \cite{du_improving_nodate,feng_dont_2024,fang2024counterfactualdebatingpresetstances}. Accuracy is calculated as the frequency of correctly predicted multiple-choice questions divided by the total question number in our dataset.

\subsubsection{Confidence}

Each time a base LLM reviews conflicting reasonings and makes new decision in the collaboration loop, it may ``concede” (change its answer) or ``insist” (not change its answer). We consider this tendency of an LLM as its intrinsic ``confidence".

We define $Q^{insist}_{\phi} \subseteq Q^{dis}$ as disagreed questions where LLM $\phi$ insists to its original answer and notice LLMs' choice is significantly affected by an group-level factor: whether there is \textbf{teammate support}. Our preliminary experiment shows a base LLM ``insists" less frequently when all other LLMs against it but more frequently if at least one LLM supports it. In other words, an LLM possess different ``confidence" level depending on teammates support level. 

To address this gap, we first classify $Q^{dis}$ by the level of teammate support:

\begin{itemize}
    \item $S^{+}$: disagreed questions where at least one teammate hold the same answer as the base model (with support)
    \item $S^{-}$: disagreed questions where none of teammate propose the same answer as the base model (without support)
\end{itemize}

Then we intersect $Q^{insist}_{\phi}$ with S\textsuperscript{+} and S\textsuperscript{-} respectively, creating a vector $\mathbf{p^{insist}_\phi}$:

\begin{equation}
\begin{aligned} 
\label{eq:P_insist}
   \mathbf{p^{insist}_\phi} &= \begin{bmatrix}\frac{|S^{+} \cap Q^{insist}_{\phi}|}{|S^{+}|}\\ \frac{|S^{-} \cap Q^{insist}_{\phi}|}{|S^{-}|}\end{bmatrix} 
\end{aligned}   
\end{equation}

Ranging from 0 to 1, larger component of $\mathbf{p^{insist}_\phi}$ indicates the LLM is more consistent to insist under specific situation S\textsuperscript{+} or S\textsuperscript{-}.

Given by we only have two levels in teammate support, we average the $p^{insist}_\phi$ vector to measure the overall confidence of $\phi$:

\begin{equation} \label{eq:confidence_formula}
\text{Confidence} = \frac{\sum\mathbf{p^{insist}_\phi}}{|\mathbf{p^{insist}_\phi}|}
\end{equation}

So far, we have built a standard confidence metric for a model $\phi$ in Equation \ref{eq:confidence_formula}. It can have maximum confidence as 1 if it insists every time in every scenario and minimal confidence as 0 if it always compromises in case of any conflict.

\subsubsection{Consistency}
Proposed by Wang \textit{et al.} \cite{wang2023selfconsistencyimproveschainthought}, self-consistency is a decoding strategy induces an LLM to generate a diverse set of reasoning paths and determine the most consistent answer. Through experiments, we observed that consistency level is a unique characteristic varying across different LLMs. We use arithmetic average metric to measure an LLM's consistency in a multiple-choice task.

To calculate the consistency, we denote $$Count(\hat{y}_{q,\phi})=|\{y_i | 1 \leq i \leq n, y_i = \hat{y}_{q, \phi}\}|$$ as the frequency of the majority vote $\hat{y}_{q,\phi}$ of a question $q$ by the LLM $\phi$, where $n$ is the parameter of self-consistency controlling the number of repeats. For example, $Count(\hat{y}_{q,\phi}) = 6$ refers to there are six votes supporting the majority vote for the question $q$ predicted by LLM $\phi$. With this, we define the consistent level of an LLM as the arithmetic average of $Count(\hat{y}_{q,\phi})$ across the dataset $Q$, adjusted by a repeat number $n$:

\begin{equation}
\label{eq:consistency}
    \text{consistency of LLM $\phi$} = \frac{\sum_{q \in Q}\frac{\text{Count}(\hat{y}_{q,\phi})}{n}}{|Q|}
\end{equation}

An LLM can have a maximum consistency level as 1 if it consistently gives out the same letter choice per multiple-choice question no matter how many times of repetition.

\section{Results and Analysis}

\subsection{Consensus Convergence}
At the initial ZS-SC-SC round, three LLMs only reach consensus on 50.82\% questions on average from $Q$ consists of 305 sample questions, but this consensus rate surges to 82.62\% after two collaboration loop iterations (Table \ref{P_con table}), which asserts our hypothesis in that collaboration mitigate the disagreements among LLMs.

\begin{table}[t]
\centering
\caption{Consensus Convergence}
\begin{tabular}{llll}
\toprule
\multicolumn{3}{c}{\textbf{Consensus Rate ($P^{con}$)}}                            \\ \midrule
                 \textbf{USMLE ($\boldsymbol{|Q|)}$} & \textbf{ZS-CoT-SC} & \textbf{Collaboration} \\ 
Step 1 (87)  & 56.32  & 86.21 \textcolor{red}{($\uparrow$ 29.89)}             \\
Step 2 (100) & 41.00  & 74.00 \textcolor{red}{($\uparrow$ 33.00)}             \\
Step 3 (118) & 55.08  & 87.29  \textcolor{red}{($\uparrow$ 32.21)}            \\ \midrule
\textbf{Average} & 50.82  & 82.62 \textcolor{red}{($\uparrow$ 31.80)}              \\ \bottomrule
\end{tabular}
\label{P_con table}
\end{table}

\subsection{Accuracy of Med42, ClinCamel, and Mixtral}
Taking the accuracy of ZS-CoT-SC as baseline, we observe every LLM has gained solid improvement after going through two collaboration loops (Table \ref{all_acc_tables}). This improvement resonates with the conclusion from prior studies regarding LLM's surprising team-working ability \cite{du_improving_nodate}. It is worth noting that our simple framework does not relay on back-and-forth communication among LLMs yet still proved to boost their individual performance.

\begin{table}[t]
\centering
\caption{Accuracy(\%) of Med42, ClinCamel, and Mixtral}
\begin{subtable}[t]{0.5\textwidth}
\captionsetup{labelfont=bf,singlelinecheck=false,
              labelsep=space,skip=2pt}
\centering
\caption{}
\begin{tabular}{@{}llllllllll@{}}
\toprule
\multicolumn{3}{c}{\textbf{Med42}}  \\ \midrule
USMLE     & ZS-CoT-SC    & Collaboration  \\
Step 1    & 81.61  & \textbf{85.06} \textcolor{red}{($\uparrow$ 3.45)}   \\
Step 2    & 64.00  & 70.00  \textcolor{red}{($\uparrow$ 6.00)} \\
Step 3    & 70.34  & 76.27  \textcolor{red}{($\uparrow$ 5.93)} \\ \midrule
\textbf{Average}  & 71.48 &76.72 \textcolor{red}{($\uparrow$ 5.24)} \\ \bottomrule
\hfill
\end{tabular}
\label{med42 accuracy_table}
\end{subtable}

\begin{subtable}[ht]{0.5\textwidth}
\captionsetup{labelfont=bf,singlelinecheck=false,
              labelsep=space,skip=2pt}
\centering	
\caption{}
\begin{tabular}{@{}lllllllllll@{}}
\toprule
\multicolumn{3}{c}{\textbf{ClinCamel}} \\ \midrule
USMLE                & ZS-CoT-SC   & Collaboration  \\
Step 1               & 65.52 & 80.46 \textcolor{red}{($\uparrow$ 14.94)} \\
Step 2               & 64.00  & 67.00 \textcolor{red}{($\uparrow$ 3.00)}\\
Step 3               & 75.42     & \textbf{78.81} \textcolor{red}{($\uparrow$ 3.39)}\\ \midrule
\textbf{Average}     & 68.85	& 75.41 \textcolor{red}{($\uparrow$ 6.56)} \\ 
\bottomrule
\hfill
\end{tabular}
\label{ClinCamel accuracy_table}
\end{subtable}

\begin{subtable}[ht]{0.5\textwidth}
\captionsetup{labelfont=bf,singlelinecheck=false,
              labelsep=space,skip=2pt}
\centering
\caption{}
\begin{tabular}{@{}lllllllllll@{}}
\toprule
\multicolumn{3}{c}{\textbf{Mixtral}} \\ \midrule
USMLE                & ZS-CoT-SC           & Collaboration \\
Step 1               & 72.41     & 79.31 \textcolor{red}{($\uparrow$ 6.90)} \\
Step 2               & 64.00  & \textbf{74.00} \textcolor{red}{($\uparrow$ 10.00)} \\
Step 3               & 77.12 & \textbf{78.81} \textcolor{red}{($\uparrow$ 1.69)} \\ \midrule
\textbf{Average}     & 71.47  & 77.38 \textcolor{red}{($\uparrow$ 5.91)} \\
\bottomrule
\hfill
\end{tabular}
\label{mixtral accuracy_table}
\end{subtable}

\label{all_acc_tables}
\end{table}

\subsection{Confidence}
In table \ref{confidence-table}, we summarize the confidence of Med42, ClinCamel, and Mixtral. For each base model, its $p^{insist}_{\phi}$ is determined by checking how many original answers at ZS-CoT-SC still remain unchanged after two collaboration loops in disregard for disagreement from other teammates. Our results suggest Med42 is the most confident model with 0.51 confidence, followed by Mixtral (0.49). ClinCamel is the most soft-minded one with only 0.23 confidence.

\begin{table}[t]
\centering
\caption{Confidence Level of Med42, ClinCamel, and Mixtral}
\centering
\begin{tabular}{@{}llllllllll@{}}

\toprule
& \multicolumn{3}{l}{\textbf{ZS-CoT-SC vs Collaboration}}    \\ \midrule
$\mathbf{P_{insist}}$ (\%) & \textbf{Med42} & \textbf{ClinCamel} & \textbf{Mixtral} \\  \midrule
W/O support ($S^{-}$)  & 45.33 & 12.99 & 42.24 \\
W/ support ($S^{+}$)   & 68.10 & 52.11 & 69.90 \\ \midrule
\textbf{Confidence} & \textbf{0.57}  & \textbf{0.33}  & \textbf{0.56}    \\ \bottomrule
\hfill
\end{tabular}
\label{confidence-table}
\end{table}

\subsection{Consistency}

In Table \ref{consistency_table}, we measure LLMs' average count of majority vote as a metric of their consistency defined in equation \ref{eq:consistency}. Due to the significant gap we observed, we separate the measurement in two situations: when the question was answered correctly and incorrectly, and use \textbf{difference} ($\boldsymbol{\Delta})$ to indicate the variation of consistency between two situations. Results from both ZS-CoT-SC and collaboration are presented.

First, all LLM non-exclusively tend to maintain high self-consistency level when their majority vote hit the ground truth but become unstable when the majority vote miss a shot.

Secondly, LLMs with greater capability and confidence such as Med42 and Mixtral also show greater $\boldsymbol{\Delta}$ while ClinCamel, which slightly underperforms the other two models in both accuracy and confidence, also has less segregation of confidence between correct and incorrect questions. 

Thirdly, compared to ZS-CoT-SC, the $\boldsymbol{\Delta}$ surges for Med42 and Mixtral after two collaboration loops but slightly decline for ClinCamel. In other words, stronger and more confident LLMs become even more affirmative with the help from teammates but it is the opposite to those weaker and easy-going LLMs -- them might fail to improve their consistency via such a collaboration work mode since ClinCamel, as an example, seems loss its perception of those questions it's able to answer correctly or not.

\begin{table}[t]
\centering
\caption{Consistency of Med42, ClinCamel, and Mixtral}

\begin{subtable}{0.5\textwidth}
\captionsetup{labelfont=bf,singlelinecheck=false,
              labelsep=space,skip=2pt}
\centering
\caption{ZS-CoT-SC}
\begin{tabular}{@{}llllllllll@{}}
\toprule
\multicolumn{4}{c}{\textbf{Average Majority Vote Count}} \\ \midrule
 & Med42   & ClinCamel  & Mixtral \\
When Correct   & 6.01 & 5.02 & 7.46 \\
When Incorrect & 1.95 & 1.59 & 1.65 \\ \midrule
\textbf{Difference} $\boldsymbol{\Delta}$ & 4.06 & 3.43 & 5.81 \\ \bottomrule
\hfill
\end{tabular}
\end{subtable}

\begin{subtable}{0.5\textwidth}
\captionsetup{labelfont=bf,singlelinecheck=false,
              labelsep=space,skip=2pt}
\centering
\caption{Collaboration}
\begin{tabular}{@{}llllllllll@{}}
\toprule
\multicolumn{4}{c}{\textbf{Average Majority Vote Count}} \\ \midrule
& Med42    & ClinCamel    & Mixtral     \\
When Correct   & 8.63 & 4.63 & 9.22     \\
When Incorrect & 0.34 & 1.79 & 0.42     \\ \midrule
\textbf{Difference} $\boldsymbol{\Delta}$ & 8.29 & 2.84 & 8.8   \\ \bottomrule
\hfill
\end{tabular}
\end{subtable}
\label{consistency_table}
\end{table}

\section{Discussion}

This paper explores whether collaboration mitigates LLMs' divergence and improves their reasoning performance in medical multiple-choice questions through a tailored workflow ICF. The results confirm both of our hypothesis and highlight some overseen facts which are discussed below.

\subsection{Confidence Plays a Role in LLM Collaboration}

Prior to our study, little research has cast light on LLM's confidence in inter-LLM collaboration and the importance confidence in such an LLM team-working setting has been understudied. 

A relative pioneer study in this field \cite{zhang_exploring_2024} use the same LLM as the backbone to create three agents, either prompted to be ``overconfident" or ``easy-going", to build up multiple LLM ``societies" and compared their accuracy. However, this prompting-based confidence customization lacks quantitative design and is not comparable between different backbone LLMs.

Feng \textit{et al.} \cite{feng_dont_2024} in their work on AbstainQA briefly allude to the idea of LLM ``confidence" in multi-LLM collaboration, stating that if an LLM, on average, changes its generations when presented with conflicting information, then it demonstrates low confidence. They suggest that LLMs, by design, should abstain from generating outputs with low confidence. However, they do not evaluate the role or effect of LLM confidence on multi-LLM collaboration. 

In our study, we do observe a pattern that suggests LLM's confidence's interplay within a team. First of all, during ZS-CoT-SC, more confident LLMs yield higher accuracy when working solely, but this trend vanished after two rounds of collaboration. However, in terms of improvement or progress in teamwork, less confident LLM benefits more from collaboration with more accuracy increase before and after collaboration. Taking both Table \ref{all_acc_tables} and Table \ref{confidence-table} into account, ClinCamel gained substantial accuracy growth of 6.56\% with only 0.23 confidence while Mixtral and Med42 gained lower accuracy improvement of 5.91\% and 5.24\% respectively, which corresponds to their higher confidence of 0.49 and 0.51. We utilize the Spearman's rank correlation \cite{doi:https://doi.org/10.1002/9781118445112.stat05964} to compute a rank-based correlation coefficient $r_s$ between accuracy improvement and confidence and obtain $r_s = -0.5$. While this moderate negative association is consistent with our observation, due to the small sample size (three LLMs) in our experiment, this test does not have statistical significance (P value = 0.67) and thus we prudentially report this result as an optional reference only. 

We discuss this ``weaker LLMs make greater progress" phenomenon from a perspective of LLM's innate assessment ability: weaker LLM, although is not good at reasoning in its own way, still possess an acute sense when judging stronger ones' reasoings and successfully align itself with the correct answer in most cases. Resonate to our finding, a study by Khan \textit{et al.} 
\cite{khan2024debatingpersuasivellmsleads} unveils non-weaker models (non-experts) achieving 32\% higher accuracy when overseeing stronger models (experts) debate each other than predicting alone. They envision this collaboration a potential solution to aligning models in the absence of ground truth.

\subsection{ICF Efficiency}

It is noteworthy that our framework is computationally light compared to related studies. 

Refer to Zhang \textit{et al.} \cite{zhang-etal-2024-exploring}, conversation-based LLM collaboration has to deal with the trade-off between conversation length and computational cost: while LLM's conformity generally grows with the conversation round increases, it demands larger consumption accordingly. However, other studies indicate that conversation length has a more dedicated effect on LLM's joint performance. Fang \textit{et al.} \cite{fang2024counterfactualdebatingpresetstances} state accuracy turns out to decrease with the increase of debate rounds. Du \textit{et al.} \cite{du_improving_nodate} also point out that as debates get longer, LLMs struggle to fully process input information.  

Unlike the mentioned LLM collaboration frameworks that requires multiple rounds of reciprocal communication between LLMs, the ``collaboration loop" in ICF (Fig.~\ref{diagram}, b) has an unidirectional circular structure that allows information flow through each base model once per round. This design greatly streamlines the pipeline and reduces the redundant encoding process of LLM inference. In addition, to prevent conversation transcripts from being too bulky to process, we set up a termination criterion $P^{con}$ to exit the collaboration loop on time when the LLMs no longer divergent. Lastly, the external summarizer LLM helps to condense the repetitive information from the lengthy concatenated paragraphs generated by self-consistency, compacting the input length in advance while preserving essential information. With fewer tokens and simpler workflow to jointly predict each medical QA question, ICF fulfills its mission well and efficiently, improving LLM insiders' accuracy while offering a straightforward way to measure their confidence. 

\subsection{Consistency Adds to LLM's Interpretability}

Refer to Table \ref{consistency_table}, in our post-hoc analysis, LLMs exhibit an evident gap in consistency with questions they predicted correctly and incorrectly. 

This interesting trait offers us a new perspective to interpret LLM's behavior: the inconsistent answers might be signals of hallucination issue. When an LLM lacks factual knowledge about the task, it tends to answer a question with false content. Since the generation of false content is relatively random, it is easily detected through self-consistency sampling as low consistency. 

Holding this assumption, we reckon it is practical to consider LLMs' consistency a potential indicator of its correctness, which could be utilized in unsupervised learning.

\section{Conclusion}
In summary, our study ensembles three LLMs in a crafted framework (ICF) for medical QA tasks, where LLMs recursively exchange reasonings on disagreed questions until reaching a consensus threshold. We discover that collaboration boosts all LLM's accuracy as well as rapidly converge their disagreement.  We propose an arithmetic metric for LLM's confidence and find that LLMs with higher confidence tend to yield higher accuracy but low-confident LLM gains more improvement in accuracy than overconfident LLMs. Last, LLMs' self-consistency is highly imbalanced among questions they predict successively and not.

\section{Acknowledgments}
This work was supported in part by the National Science Foundation under the Grants IIS-1741306 and IIS-2235548, and by the Department of Defense under the Grant DoD W91XWH-05-1-023. This material is based upon work supported by (while serving at) the National Science Foundation. Any opinions, findings, and conclusions or recommendations expressed in this material are those of the author(s) and do not necessarily reflect the views of the National Science Foundation. In addition, we would like to express our sincere gratitude to Mary M. Lucas at Drexel University College of Computing and Informatics for her assistance in reviewing, verifying, and providing insightful feedback to this work. 

\bibliographystyle{ieeetran}
\bibliography{IEEEabrv,Bibs}
\end{document}